# BB8: A Scalable, Accurate, Robust to Partial Occlusion Method for Predicting the 3D Poses of Challenging Objects without Using Depth

Mahdi Rad[1]    Vincent Lepetit[1, 2]

[1]Institute for Computer Graphics and Vision, Graz University of Technology, Austria

[2] Laboratoire Bordelais de Recherche en Informatique, Université de Bordeaux, Bordeaux, France

{rad, lepetit}@icg.tugraz.at

## Abstract

*We introduce a novel method for 3D object detection and pose estimation from color images only. We first use segmentation to detect the objects of interest in 2D even in presence of partial occlusions and cluttered background. By contrast with recent patch-based methods, we rely on a "holistic" approach: We apply to the detected objects a Convolutional Neural Network (CNN) trained to predict their 3D poses in the form of 2D projections of the corners of their 3D bounding boxes. This, however, is not sufficient for handling objects from the recent T-LESS dataset: These objects exhibit an axis of rotational symmetry, and the similarity of two images of such an object under two different poses makes training the CNN challenging. We solve this problem by restricting the range of poses used for training, and by introducing a classifier to identify the range of a pose at run-time before estimating it. We also use an optional additional step that refines the predicted poses. We improve the state-of-the-art on the LINEMOD dataset from 73.7% [2] to 89.3% of correctly registered RGB frames. We are also the first to report results on the Occlusion dataset [1] using color images only. We obtain 54% of frames passing the Pose 6D criterion on average on several sequences of the T-LESS dataset, compared to the 67% of the state-of-the-art [10] on the same sequences which uses both color and depth. The full approach is also scalable, as a single network can be trained for multiple objects simultaneously.*

## 1. Introduction

3D pose estimation of object instances has recently become a popular problem again, because of its application in robotics, virtual and augmented reality. Many recent approaches rely on depth maps, sometimes in conjunction with color images [5, 4, 9, 14, 7, 1, 21, 13, 3, 10]. However, it is not always possible to use depth cameras, as they fail outdoor or on specular objects. In addition, they drain the batteries of mobile devices, being an active sensor.

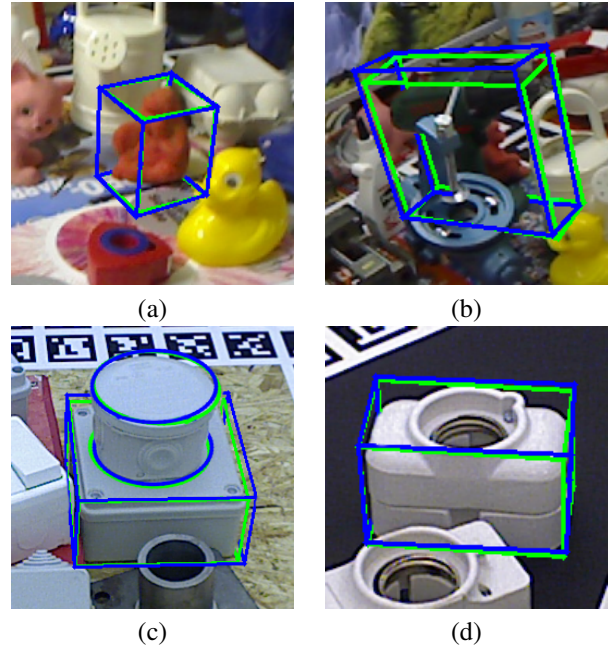

Figure 1. Zooms on estimated poses for (a) the Ape of the LINEMOD dataset [7], (b) the Driller of the Occlusion dataset [1], (c) and (d) three objects of the T-LESS [10] dataset. The green bounding boxes correspond to the ground truth poses, and the blue bounding boxes to the poses estimated with our method. The two boxes often overlap almost perfectly, showing the accuracy of our estimated poses. The parts of the bounding boxes occluded by the object were removed using the object mask rendered from our estimated pose. In (b), we can still obtain a good pose despite the large occlusion by the bench vise. In (c) and (d), we also obtain very good estimates despite large occlusions, the similarities between the objects, and the fact that the symmetries challenge the learning algorithms.

It is therefore desirable to rely only on color images for 3D pose estimation, even if it is more challenging. Recent methods [1, 13, 2] work by identifying the 'object coordinates' of the pixels, which are the pixels' 3D coordinates in a coordinate system related to the object [19]. The object 3D pose can then be estimated using a P$n$P algorithm from these 2D-3D correspondences. [3] obtain similar correspondences by associating some pixels in selected parts of the object with virtual 3D points. However, obtaining these 2D-3D correspondences from local patches is difficult and the output is typically very noisy for these methods. A robust optimization is then needed to estimate the pose.

In this paper, we argue for a "holistic" approach, in the sense that we predict the pose of an object directly from its appearance, instead of identifying its individual surface points. As we will show, this approach provides significantly better results.

We first detect the target objects in 2D. We show that using object segmentation performs better for this task compared to a standard sliding window detector, in particular in presence of partial occlusion. We then apply a CNN to predict the 3D pose of the detected objects. While the predicted 3D pose can be represented directly by a translation and a rotation, we achieve better accuracy by using a representation similar to the one used in [3] for object parts: We predict the 2D projections of the corners of the object's bounding box, and compute the 3D pose from these 2D-3D correspondences with a P$n$P algorithm. Compared to the object coordinate approaches the predictions are typically outlier-free, and no robust estimation is thus needed. Compared to the direct prediction of the pose, this also avoids the need for a meta-parameter to balance the translation and rotation terms.

Unfortunately, this simple approach performs badly on the recent and challenging T-LESS dataset. This dataset is made of manufactured objects that are not only similar to each other, but also have one axis of rotational symmetry. For example, the squared box of Fig. 1(c) has an angle of symmetry of $90°$ and the other object has an angle of symmetry of $0°$ since it is an object of revolution; Object #5 in Fig. 1(d) is not perfectly symmetrical but only because of the small screw on the top face.

The approach described above fails on these objects because it tries to learn a mapping from the image space to the pose space. Since two images of a symmetrical object under two different poses look identical, the image-pose correspondence is in fact a one-to-many relationship. This issue is actually not restricted to our approach. For example, [2], which relies on object coordinates, does not provide results on the Bowl object of the LINEMOD dataset, an object with an axis of symmetry: It is not clear which coordinates should be assigned to the 3D points of this object, as all the points on a circle orthogonal to the axis of symmetry have the same appearance.

To solve this problem, we train the method described above using images of the object under rotation in a restricted range, such that the training set does not contain ambiguous images. In order to recover the object pose under a larger range of rotation, we train a classifier to tell under which range the object rotation is. Again, this is easy to do with a "holistic" approach, and this classifier takes an image of the entire object as input. As we will explain in more details, we can then always use the CNN trained on the restricted range to estimate any pose. In addition, we will show how to adapt this idea to handle "approximatively symmetrical" objects like Object #5. This approach allows us to obtain good performance on the T-LESS dataset.

Finally, we show that we can add an optional last step to refine the pose estimates by using the "feedback loop" proposed in [17] for hand detection in depth images: We train a network to improve the prediction of the 2D projections by comparing the input image and a rendering of the object for the initial pose estimate. This allows us to improve even more our results on the LINEMOD and Occlusion datasets.

Our full approach, which we call BB8, for the 8 corners of the bounding box, is also very fast, as it only requires to apply Deep Networks to the input image a few times. In the remainder of the paper, we first discuss related work, describe our approach, and compare it against the state-of-the-art on the three available datasets.

## 2. Related Work

The literature on 3D object detection is very large, thus we will focus only on recent works. Keypoint-based methods [16, 23] were popular for a long time and perform well but only on very textured objects. The apparition of inexpensive 3D cameras favored the development of methods suitable for untextured objects: [5, 9] rely on depth data only and use votes from pairs of 3D points and their normals to detect 3D objects. [14] uses a decision tree applied to RGB-D images to simultaneously recognize the objects and predict their poses. [7, 24] consider a template-based representation computed from RGB-D or RGB data, which allows for large scale detection [11]. However, this template approach is sensitive to partial occlusions.

To tackle clutter and partial occlusions, [1] and [21] rely on local patches recognition performed with Random Forests. In particular, [1] considers '3D object coordinates': A Random Forest is trained to predict the 3D location in the object coordinate system of each image location. The prediction of this forest is integrated in an energy function together with a term that compares the depth map with a rendering of the object and a term that penalizes pixels that lie on the object rendering but predicted by the forest to not be an object point. This energy function is optimized by a RANSAC procedure. [13] replaces this energy function by an energy computed from the output of a CNN trained to compare observed image features and features computed from a 3D rendering of the potentially detected object. This makes the approach very robust to partial occlusions.

These works, however, are designed for RGB-D data. [2] extends this work and relies on RGB data only, as we do. They use Auto-Context [22] to obtain better predictions from the Random Forests, estimate a distribute over the object coordinates to handle the prediction uncertainties better, and propose a more sophisticated RANSAC-like method that scales with the number of objects. This results in an efficient and accurate method, however, robustness to partial occlusions are not demonstrated.

[3] is related to [1, 21, 2] but focuses on providing sparse 2D-3D correspondences from reliable object parts. Unfortunately, it provides results on its own dataset only, not on more broadly available datasets.

Like us, [12] relies on a CNN to directly predict a 3D pose, but in the form of a translation and a rotation. It considers camera relocalisation in urban environment rather than 3D object detection, and uses the full image as input to the CNN. By predicting the 2D projections of the corners of the bounding box, we avoid the need for a meta-parameter to balance the position and orientation errors. As shown in our experiments, the pose appears to be more accurate when predicted in this form. Intuitively, this should not be surprising, as predicting 2D locations from a color images seems easier than predicting a 3D translation and a quaternion, for example.

[6] also uses a CNN to predict the 3D pose of generic objects but from RGB-D data. It first segments the objects of interest to avoid the influence of clutter. We tried segmenting the objects before predicting the pose as well, however, this performed poorly on the LINEMOD dataset, because the segmented silhouttes were not very accurate, even with state-of-the-art segmentation methods.

In summary, our method appears to be one of the first to deal with RGB data only to detect 3D objects and estimate their poses on recent datasets. As we will show in the experiments, it outperforms the accuracy of the state-of-the-art [2] by a large margin.

## 3. Proposed Approach

In our approach, we first find the objects in 2D, we obtain a first estimate of the 3D poses, including objects with a rotational symmetry, and we finally refine the initial pose estimates. We describe each step in this section.

### 3.1. Localizing the Objects in 2D

We first identify the 2D centers of the objects of interest in the input images. We could use a standard 2D object detector, but we developed an approach based on segmentation that resulted in better performance as it can provide accurate locations even under partial occlusions. Compared to our initial tests using a sliding window, this ap-proach improved our 2D detection results from about 75% to 98.8% correct detection rate based on a IoU of 0.5. We only need a low resolution segmentation and thus do not need a hourglass-shaped architecture [15], which makes our segmentation more efficient.

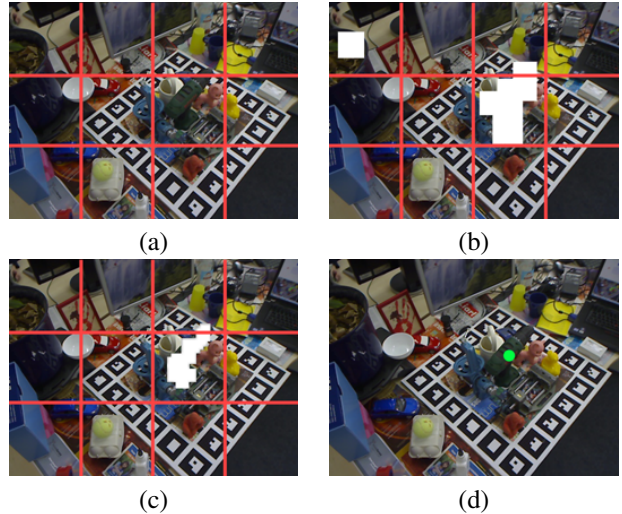

Figure 2. Object localization using our segmentation approach: (a) The input image is resized to $512 \times 384$ and split into regions of size $128 \times 128$. (b) Each region is first segmented into a binary mask of $8 \times 8$ for each possible object $o$. (c) Only the largest component is kept if several components are present, the active locations are segmented more finely. (d) The centroid of the final segmentation is used as the 2D object center.

As shown in Fig. 2, our approach performs a two-level coarse-to-fine object segmentation. For each level, we train a single network for all the objects. The first network is obtained by replacing the last layer of VGG [20] by a fully connected layer with the required number of output required by each step, and fine-tune it. The second network has a simple, *ad hoc* architecture.

More exactly, the first network is trained to provide a very low resolution binary segmentation of the objects given an image region $J$ of size $128 \times 128$ by minimizing the following objective function:

$$\sum_{(J,S,o)\in\mathcal{T}_s} \|(f_\phi^1(J))[o] - S\|^2 \,, \qquad (1)$$

where $\mathcal{T}_s$ is a training set made of image regions $J$, and the corresponding segmentations $S$ for object $o$, $(f_\phi^1(J))[o]$ is the output of network $f_\phi^1$ for region $J$ and object $o$. $\phi$ denotes the network's parameters, optimized during training. For the LINEMOD and Occlusion datasets, there is at most one object for a given region $J$, but more objects can be present for the T-LESS dataset. At run-time, to get the segmentations, we compute:

$$s_{1,o}(J) = (f_\phi^1(J))[o] > \tau_1 \,, \qquad (2)$$

where $s_{1,o}$ is a $8 \times 8$ binary segmentation of $J$ for object $o$, and $\tau_1$ is a threshold used to binarize the network's output. To obtain a binary segmentation for the full input image, we split this image into regions and compute the $s_{1,o}$ for each region.

This gives us one binary segmentation $S_{1,o}$ for the full input image, and each possible object. This usually results in a single connected component per visible object; if several components are present, we keep only the largest one for each object. If the largest component in a segmentation $S_{1,o}$ is small, object $o$ is likely not visible. For the remaining object(s), we refine the shape of the largest component by applying a second network to each $16 \times 16$ image patch $P$ that corresponds to an active location in $S_1$:

$$s_{2,o}(P) = (f_\psi^2(P))[o] > \tau_2 \,, \quad (3)$$

using notations similar to the ones in Eq. (2). Since the input to $f_\psi^2(P)$ has a low resolution, we do not need a complex network such as VGG [20], and we use a much simpler architecture with 2 convolutional layers and 2 pooling layers. We finally obtain a segmentation $S_{2,o}$ with resolution $64 \times 48$ for the full input image and each visible object $o$. We therefore get the identities $o$ of the visible object(s), and for these objects, we use the segmentation centroids as their 2D centers, to compute the 3D poses of the objects as described below.

### 3.2. Predicting the 3D Pose

We predict the 3D pose of an object by applying a Deep Network to an image window $W$ centered on the 2D object center estimated as described in the previous section. As for the segmentation, we use VGG [20] as a basis for this network. This allows us to handle all the objects of the target dataset with a single network.

It is possible to directly predict the pose in the form of a 3-vector and an exponential map for example, as in [12]. However, a more accurate approach was proposed in [3] for predicting the poses of object parts. To apply it here, we minimize the following cost function over the parameters $\Theta$ of network $g_\Theta$:

$$\sum_{(W,\mathbf{e},\mathbf{t},o)\in\mathcal{T}} \sum_i \|\text{Proj}_{\mathbf{e},\mathbf{t}}(\mathbf{M}_i^o) - m_i((g_\Theta(W))[o])\|^2 \,, \quad (4)$$

where $\mathcal{T}$ is a training set made of image windows $W$ containing object $o$ under a pose defined by an exponential map $\mathbf{e}$ and a 3-vector $\mathbf{t}$. The $\mathbf{M}_i^o$ are the 3D coordinates of the corners of the bounding box of object $o$ in the object coordinate system. $\text{Proj}_{\mathbf{e},\mathbf{t}}(\mathbf{M})$ projects the 3D point $\mathbf{M}$ on the image from the pose defined by $\mathbf{e}$ and $\mathbf{t}$. $m_i((g_\Theta(W))[o])$ returns the two components of the output of $g_\Theta$ corresponding to the predicted 2D coordinates of the $i$-th corner for object $o$.

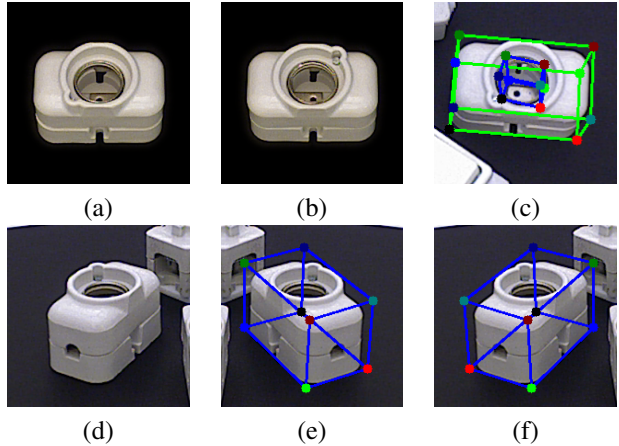

Figure 3. Handling Objects with a symmetry of rotation: Object #5 of T-LESS has an angle of symmetry $\alpha$ of $180°$, if we ignore the small screw and electrical contact. If we restrict the range of poses in the training set between $0°$ (a) and $180°$ (b), pose estimation still fails for test samples with an angle of rotation close to $0°$ modulo $180°$ (c). Our solution is to restrict the range during training to be between $0°$ and $90°$. We use a classifier to detect if the pose in an input image is between $90°$ and $180°$. If this is the case (d), we mirror the input image (e), and mirror back the predicted projections for the corners (f).

At run-time, the segmentation gives the identity and the 2D locations of the visible object(s) $o$. The 3D pose can then be estimated for the correspondences between the 3D points $\mathbf{M}_i^o$ and the predicted $m_i((g_\Theta(W))[o])$ using a PnP algorithm. Other 3D points could be used here, however, the corners of the bounding box are a natural choice as they frame the object and are well spread in space [1].

### 3.3. Handling Objects with an Axis of Symmetry

If we apply the method described so far to the T-LESS dataset, the performances are significantly lower than the performances on the LINEMOD dataset. As mentioned in the introduction, this is because training images $W$ in Eq. (4) for the objects of this dataset can be identical while having very different expected predictions $\text{Proj}_{\mathbf{e},\mathbf{t}}(\mathbf{M}_i^o)$, because of the rotational symmetry of the objects.

We first remark that for an object with an angle of symmetry $\alpha$, its 3D rotation around its axis of symmetry can be defined only modulo $\alpha$, not $2\pi$. For an object with an angle of symmetry $\alpha$, we can therefore restrict the poses used for training to the poses where the angle of rotation around the symmetry axis is within the range $[0; \alpha[$, to avoid the ambiguity between images. However, this solves our problem only partially: Images at one extremity of this range of poses and the images at the other extremity, while not identical, still look very similar. As a result, for input images with an angle of rotation close to 0 modulo $\alpha$, the pose prediction can still be very bad, as illustrated in Fig. 3.

---
[1]The bounding boxes shown in the figures of this paper were obtained by projecting the 3D bounding box given the recovered poses, not directly from the output of $g_\Theta$.

To explain our solution, let us first denote by $\beta$ the rotation angle, and introduce the intervals $r_1 = [0; \alpha/2[$ and $r_2 = [\alpha/2; \alpha[$. To avoid ambiguity, we restrict $\beta$ to be in $r_1$ for the training images used in the optimization problem of Eq. (4). The drawback is of course that, without doing anything else, we would not be able to estimate the poses when $\beta$ is in $r_2$.

We therefore introduce a CNN classifier $k(\cdot)$ to predict at run-time if $\beta$ is in $r_1$ or $r_2$: If $\beta$ is in $r_1$, we can estimate the pose as before; If $\beta$ is in $r_2$, one option would be to apply another $g_\Theta(\cdot)$ network trained for this range.

However, it is actually possible to use the same network $g_\Theta(\cdot)$ for both $r_1$ and $r_2$, as follows. If the classifier predicts that $\beta$ in in $r_2$, we mirror the input image $W$: As illustrated in Fig. 3(e), the object appears in the mirror image with a rotation angle equal to $\alpha - \beta$, which is in $r_1$. Therefore we can apply $g_\Theta(\cdot)$ to the mirrored $W$. To obtain the correct pose, we finally mirror back the projections of the corners predicted by $g_\Theta(\cdot)$. We currently consider the case where the axis of symmetry is more or less vertical in the image, and mirror the image from left to right. When the axis is closer to be horizontal, we should mirror the image from top to bottom.

Objects of revolution are a special and simpler case: since their angle of symmetry is 0°, we predict their poses under the same angle of rotation. For training the pose predictor $g_\Theta(\cdot)$, we use the original training images with angles of rotation in $r_1$, and mirror the training images with angles of rotation in $r_2$.

**Handling Objects that are 'Not Exactly Symmetrical'**
As mentioned in the introduction, some objects of the T-LESS dataset are only approximately symmetrical, such as Object #5 in Fig. 1(d). The small details that make the object not perfectly symmetrical, however, do not help the optimization problem of Eq. (4), but we would still like to predict the pose of this object.

In the case of Object #5, we consider 4 regions instead of 2: $r_1 = [0; \pi/2[$, $r_1 = [\pi/2; \pi[$, $r_3 = [\pi; 3\pi/2[$, and $r_4 = [3\pi/2; 2\pi[$, and we train the classifier $k(\cdot)$ to predict in which of these four regions the angle of rotation $\beta$ is. If $\beta \in r_2$ or $\beta \in r_4$, we mirror the image before computing the pose as before. Then, if $\beta \in r_3$ or $\beta \in r_4$, we still have to add $\pi$ to the angle of rotation of the recovered pose to get an angle between 0 and $2\pi$.

### 3.4. Refining the Pose

We also introduce an optional additional stage to improve the accuracy of the pose estimates inspired by [17].

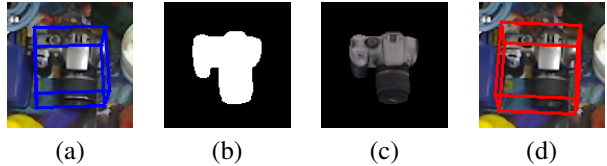

Figure 4. Refining the pose. Given a first pose estimate, shown by the blue bounding box (a), we generate a binary mask (b) or a color rendering (c) of the object. Given the input image and this mask or rendering, we can predict an update that improves the object pose, shown by the red bounding box (d).

As illustrated in Fig. 4, we train another CNN that predicts an update to improve the pose. Because this CNN takes 4 or 6 channels as input, it is not clear how we can use VGG, as we did for the previously introduced networks, and we use here one CNN per object. However, this stage is optional, and without it, we already outperform the-state-of-the-art. The first image is the image window $W$ as for $g_\Theta(\cdot)$. The second image depends on the current estimate of the pose: While [17] generates a depth map with a deep network, we render (using OpenGL) either a binary mask or a color rendering of the target object as seen from this current estimate. More formally we train this CNN by minimizing:

$$\sum_{(W,\mathbf{e},\mathbf{t})\in\mathcal{T}} \sum_{(\hat{\mathbf{e}},\hat{\mathbf{t}})\in\mathcal{N}(\mathbf{e},\mathbf{t})} \sum_i \|\text{Proj}_{\mathbf{e},\mathbf{t}}(\mathbf{M}_i^o) - \text{Proj}_{\hat{\mathbf{e}},\hat{\mathbf{t}}}(\mathbf{M}_i^o) - m_i(h_\mu(W, \text{Render}(\hat{\mathbf{e}},\hat{\mathbf{t}})))\|^2, \tag{5}$$

where $h_\mu$ denotes the CNN, $\mu$ its parameters; $\mathcal{N}(\mathbf{e},\mathbf{t})$ is a set of poses sampled around pose $(\mathbf{e},\mathbf{t})$, and $\text{Render}(\mathbf{e},\mathbf{t})$ a function that returns a binary mask, or a color rendering, of the target object seen from pose $(\mathbf{e},\mathbf{t})$.

At run-time, given a current estimate of the object pose represented by the projections of the corners $\hat{\mathbf{v}} = [\ldots \hat{\mathbf{m}}_i^\top \ldots]^\top$, and the corresponding parameterisation $(\hat{\mathbf{e}},\hat{\mathbf{t}})$, we can update this estimate by invoking $h_\mu(\cdot)$:

$$\hat{\mathbf{v}} \leftarrow \hat{\mathbf{v}} + h_\mu(W, \text{Render}(\hat{\mathbf{e}},\hat{\mathbf{t}})). \tag{6}$$

### 3.5. Generating Training Images

In Section 4, we will compare our method to the state-of-the art for 3D object detection in color images [2], and like them, for each of 15 objects of the LINEMOD dataset, we use 15% of the images for training and use the rest for testing. The training images are selected as in [2], such that relative orientation between them should be larger than a threshold. We also tried a random selection, and there was only a slight drop in performance, for some objects only. The selection method thus does not seem critical. The T-LESS dataset provides regularly sampled training images.

As shown in Fig. 5, we also use a similar method as [2] to augment the training set: We extract the objects' silhouettes from these images, which can be done as the ground truth poses and the objects' 3D models are available. Note that this means the results are not influenced by the scene context, which makes the pose estimation more difficult.

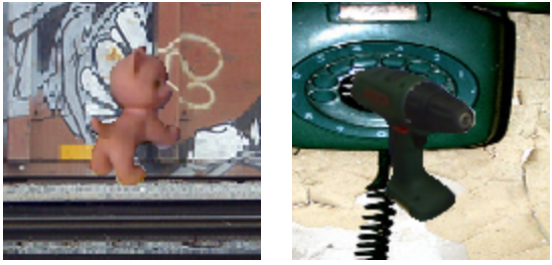

Figure 5. Two generated training images for different objects from the LINEMOD dataset [7]. The object is shifted from the center to handle the inaccuracy of the detection method, and the background is random to make sure that the network $g_\Theta$ cannot exploit the context specific to the dataset.

| Sequence | Direct | BB | Mask Ref. | RGB Ref. |
|---|---|---|---|---|
| Ape (*) | 91.2 | 96.2 | 97.5 | **97.7** |
| Bench Vise | 61.3 | 80.2 | 90.1 | **91.5** |
| Camera | 43.1 | 82.8 | 82.5 | **86.3** |
| Can | 62.5 | 85.8 | 90.2 | **91.5** |
| Cat (*) | 93.1 | 97.2 | **98.6** | 98.6 |
| Driller (*) | 46.5 | 77.6 | 83.4 | **83.6** |
| Duck | 67.9 | 84.6 | 94.0 | **94.1** |
| Egg Box | 68.2 | 90.1 | 92.0 | **93.2** |
| Glue | 69.3 | 93.5 | 94.2 | **95.8** |
| Hole Puncher | 78.2 | 91.7 | 95.2 | **97.4** |
| Iron | 64.5 | 79.0 | 79.5 | **85.0** |
| Lamp | 50.4 | 79.9 | **83.6** | 83.5 |
| Phone | 46.9 | 80.0 | 85.6 | **88.9** |
| average | 64.9 | 85.4 | 89.7 | **91.3** |

Table 1. Evaluation using the 2D Projections metric of using the 2D projections of the bounding box ('BB'), compared to the direct prediction of the pose ('Direct'), and of the refinement methods. For this evaluation, we used the ground truth 2D object center to avoid the influence of the detection. For the objects marked with a (*), we optimize the value of the weight balancing the rotation and translation terms on the test set, giving an advantage to the 'Direct' pose method. For the other objects, we used the value that is optimal for both the Ape and the Driller.

To be robust to clutter and scale changes, we scale the segmented objects by a factor of $s \in [0.8, 1.2]$, and change the background by a patch extracted from a randomly picked image from the ImageNet dataset [18]. Moreover, the object is shifted by some pixels from the center of the image window in both $x$ and $y$ directions. This helps us to handle small object localization errors made during detection.

## 4. Experiments

In this section, we present and discuss the results of our evaluation. We first describe the three evaluation metrics used in the literature and in this paper. We evaluate our method on all the possible datasets with color images for instance 3D detection and pose estimation we are aware of: the LINEMOD [7], Occlusion [1], and T-LESS [10] datasets.

### 4.1. Evaluation Metrics

As in [2], we use the percentage of correctly predicted poses for each sequence and each object, where a pose is considered correct if it passes the tests presented below.

**2D Projections [2]** This is a metric suited for applications such as augmented reality. A pose is considered correct if the average of the 2D distances between the projections of the object's vertices from the estimated pose and the ground truth pose is less than 5 pixels.

**6D Pose [8]** With this metric, a pose is considered correct if the average of the 3D distances between the transformed of the object's vertices

$$\frac{1}{|\mathcal{V}|} \sum_{\mathbf{M} \in \mathcal{V}} \|\text{Tr}_{\hat{\mathbf{e}}, \hat{\mathbf{t}}}(\mathbf{M}) - \text{Tr}_{\bar{\mathbf{e}}, \bar{\mathbf{t}}}(\mathbf{M})\|_2 \quad (7)$$

is less than 10% of the object's diameter. $\mathcal{V}$ is the set of the object's vertices, $(\hat{\mathbf{e}}, \hat{\mathbf{t}})$ the estimated pose and $(\bar{\mathbf{e}}, \bar{\mathbf{t}})$ the ground truth pose, and $\text{Tr}_{\mathbf{e}, \mathbf{t}}(\cdot)$ a rigid transformation by rotation $\mathbf{e}$, translation $\mathbf{t}$. For the objects with ambigious poses due to symmetries, [8] replaces this measure by:

$$\frac{1}{|\mathcal{V}|} \sum_{\mathbf{M}_1 \in \mathcal{V}} \min_{\mathbf{M}_2 \in \mathcal{V}} \|\text{Tr}_{\hat{\mathbf{e}}, \hat{\mathbf{t}}}(\mathbf{M}_1) - \text{Tr}_{\bar{\mathbf{e}}, \bar{\mathbf{t}}}(\mathbf{M}_2)\|_2 \, . \quad (8)$$

**5cm 5° Metric [19]** With this metric, a pose is considered correct if the translation and rotation errors are below 5cm and 5° respectively.

### 4.2. Contributions of the Different Steps

The columns 'BB', 'Mask Ref.', and 'RGB Ref.' of Table 1 compare the results of our method before and after two iterations of refinement, using either a binary mask or a color rendering. For this evaluation, we used the ground truth 2D object center to avoid the influence of the detection. Using refinement improves the results on average by 4.5% and 6.3% for the mask and color rendering respectively. Using a color rendering systematically yields the best results, but using the binary mask yields already a significant improvement, showing that an untextured model can be used.

| Metric | 2D Projection | | | 6D Pose | | 5cm 5° | |
|---|---|---|---|---|---|---|---|
| Sequence | [2] | w/o | w/Ref. | [2] | w/Ref. | [2] | w/Ref. |
| Ape | 85.2 | 95.3 | **96.6** | 33.2 | **40.4** | 34.4 | **80.2** |
| Bench Vi. | 67.9 | 80.0 | **90.1** | 64.8 | **91.8** | 40.6 | **81.5** |
| Camera | 58.7 | 80.9 | **86.0** | 38.4 | **55.7** | 30.5 | **60.0** |
| Can | 70.8 | 84.1 | **91.2** | 62.9 | **64.1** | 48.4 | **76.8** |
| Cat | 84.2 | 97.0 | **98.8** | 42.7 | **62.6** | 34.6 | **79.9** |
| Driller | 73.9 | 74.1 | **80.9** | 61.9 | **74.4** | 54.5 | **69.6** |
| Duck | 73.1 | 81.2 | **92.2** | 30.2 | **44.3** | 22.0 | **53.2** |
| Egg Box | 83.1 | 87.9 | **91.0** | 49.9 | **57.8** | 57.1 | **81.3** |
| Glue | 74.2 | 89.0 | **92.3** | 31.2 | **41.2** | 23.6 | **54.0** |
| Hole P. | 78.9 | 90.5 | **95.3** | 52.8 | **67.2** | 47.3 | **73.1** |
| Iron | 83.6 | 78.9 | **84.8** | 80.0 | **84.7** | 58.7 | **61.1** |
| Lamp | 64.0 | 74.4 | **75.8** | 67.0 | **76.5** | 49.3 | **67.5** |
| Phone | 60.6 | 77.6 | **85.3** | 38.1 | **54.0** | 26.8 | **58.6** |
| average | 73.7 | 83.9 | **89.3** | 50.2 | **62.7** | 40.6 | **69.0** |
| Bowl | - | 97.0 | **98.9** | - | **60.0** | - | **90.9** |
| Cup | - | 93.4 | **94.8** | - | **45.6** | - | **58.4** |

Table 2. Comparison between [2] and our method without and with RGB Refinement using our segmentation-based method to obtain the 2D object centers on the LINEMOD dataset. [2] does not provide results for the Bowl and the Cup, hence for the sake of comparison the average is taken over the first 13 objects.

### 4.3. The LINEMOD Dataset: Comparison with [2]

Table 2 compares our BB8 method with and without RGB refinement against the one presented in [2] on the LINEMOD dataset. Because of lack of space, we provide the results without refinement only for the 2D Projection metric, however, the results for the other metrics are comparable. For this evaluation, we used the results of our detection method presented in Section 3.1, *not* the ground truth 2D object center. Our method outperforms [2] by a large margin: 15.6% for 2D Projection, 12.6% for 6D Pose and 28.4% for the $5cm\ 5°$ metric.

Fig. 7 shows qualitative results for our method on this dataset. For most of the images, the two bounding boxes, for the ground truth pose and for the pose we estimate, overlap almost perfectly.

### 4.4. The Occlusion Dataset: Robustness to Partial Occlusions

The Occlusion dataset was created by [1] from the LINEMOD dataset. The partial occlusions make it significantly more difficult, and to the best of our knowledge, the only published results use both color and depth data. [2] provide results using only color images, but limited to 2D detection, not 3D pose estimation.

We only use images from the LINEMOD dataset to generate our training images by using the approach explained in Section 3.5, except that we also randomly superimpose objects extracted from the other sequences to the target object to be robust to occlusions. We *do not* use any image of the test sequence to avoid having occlusions similar to the ones presented in the test sequence.

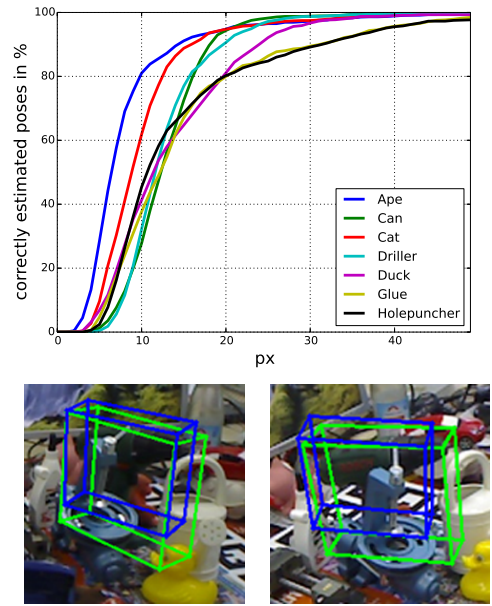

Figure 6. Performance on the Occlusion dataset [1]. Top: Percentages of correctly estimated poses as a function of the distance threshold of the 2D projections metric for 'BB8' on the Occlusion dataset. For a 15px threshold, about 80% of the frames are correctly registered, and about 90% for a 20px threshold. Bottom: Two registered frames for the Driller with a 15px and 20px error respectively.

Although all the poses in the test sets are not visible in the training sequences, we can estimate accurate poses with a 2D Projection error lower than 15px for about 80% of the frames for these seven objects. We do not report the performance of our method for the Eggbox, as more than 70% of close poses are not seen in the training sequence. Some qualitative results are shown in the second row of Fig. 7. To the best of our knowledge, we are the first to present results on this dataset using color images only.

### 4.5. The T-LESS Dataset: Handling Objects with an Axis of Symmetry

The test sequences of the T-LESS dataset are very challenging, with sometimes multiple instances of the same objects and a high amount of clutter and occlusion. We considered only Scenes #1, #2, #4, #5, and #7 in our experiments. It is also difficult to compare against the only published work on T-LESS [10], as it provides the 6D pose metric averaged per object or per scene, computed using RGB-D data, while, to the best of our knowledge, we are the first to report results on the T-LESS dataset using RGB images only. Similarly to [10], we evaluate the poses with more than 10% of the object surface visible in the ground truth poses. As shown in Table 3, the 6D Pose average per scene with our method is 54%. The object 3D orientation and translation along the $x$ and $y$ axes of the camera are typically very well estimated, and most of the error is along the $z$ axis, which should not be surprising for a method using color images only.

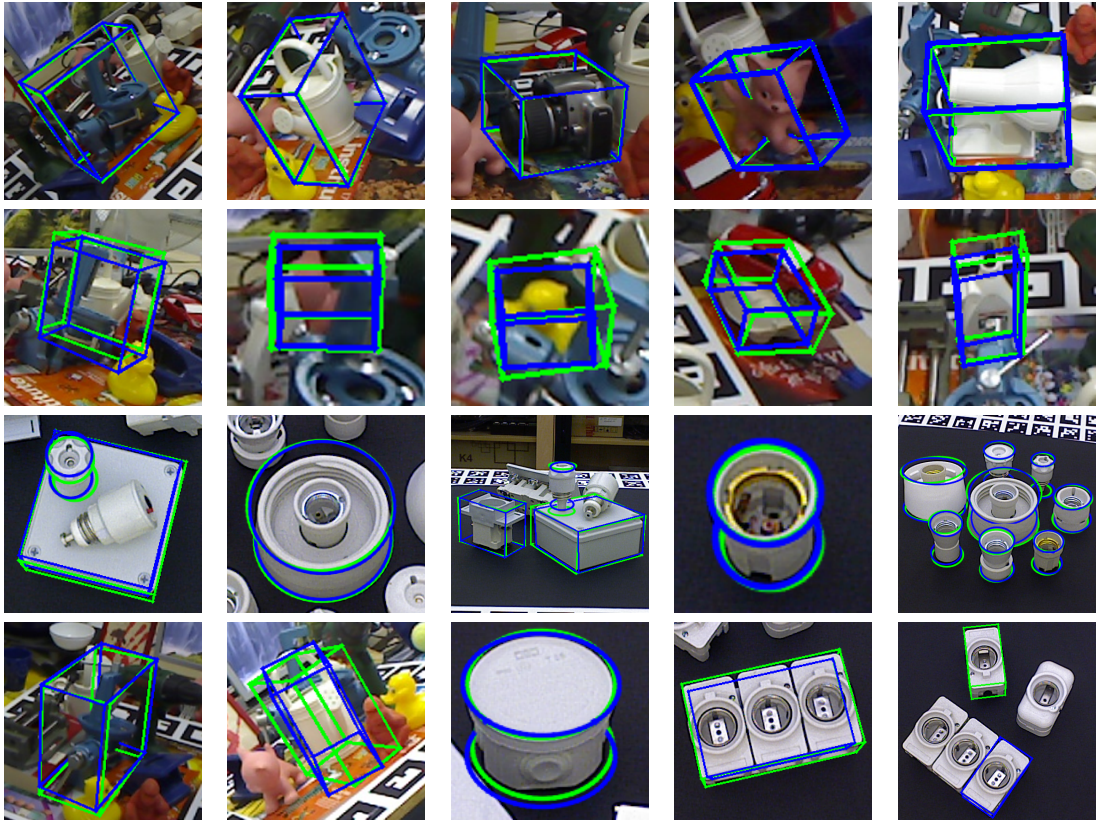

Figure 7. Some qualitative results. First row: LINEMOD dataset; Second row: Occlusion dataset; Third row: T-LESS dataset (for objects of revolution, we represent the pose with a cylinder rather than a box); Last row: Some failure cases. From left to right: An example of a pose rejected by the 2D Projections metric, a failure due to the lack of corresponding poses in the training set, two examples from T-LESS rejected by the 6D pose metric, and one failure due to the fact that some objects are made of several instances of another object.

| Scene ID: [Obj. IDs] | 6D Pose | Average |
|---|---|---|
| 1: [2, 30] | 50.8, 55.4 | 53.1 |
| 2: [5, 6] | 56.5, 55.6 | 56.1 |
| 4: [5, 26, 28] | 68.7, 53.3, 40.6 | 54.3 |
| 5: [1, 10, 27] | 39.6, 69.9, 50.1 | 53.2 |
| 7: [1, 3, 13, 14, ... | 42.0, 61.7, 64.5, 40.7, ... | |
| 7: ... 15, 16, 17, 18] | ...39.7, 45.7, 50.2, 83.7 | 53.5 |

Table 3. Our quantitative results on T-LESS [10]. Most of the errors are along the $z$ axis of the camera, as we rely on color images.

### 4.6. Computation Times

Our implementation takes 140 ms for the segmentation, 130 ms for the pose prediction, and 21 ms for each refinement iteration, on an Intel Core i7-5820K 3.30 GHz desktop with a GeForce TITAN X. If there is only one object of interest, we can replace VGG by a specific network with a simpler architecture, the computation times then become 20 ms for the segmentation and 12 ms for the pose prediction, with similar accuracy.

### 5. Conclusion

Our "holistic" approach, made possible by the remarkable abilities of Deep Networks for regression, allowed us to significantly advance the state-of-the-art on 3D pose estimation from color images, even on challenging objects from the T-LESS dataset.

**Acknowledgment:** This work was funded by the Christian Doppler Laboratory for Semantic 3D Computer Vision.

# References

[1] E. Brachmann, A. Krull, F. Michel, S. Gumhold, J. Shotton, and C. Rother. Learning 6D Object Pose Estimation Using 3D Object Coordinates. In *ECCV*, 2014. 1, 2, 3, 6, 7

[2] E. Brachmann, F. Michel, A. Krull, M. M. Yang, S. Gumhold, and C. Rother. Uncertainty-Driven 6D Pose Estimation of Objects and Scenes from a Single RGB Image. In *CVPR*, 2016. 1, 2, 3, 5, 6, 7

[3] A. Crivellaro, M. Rad, Y. Verdie, K. Yi, P. Fua, and V. Lepetit. A Novel Representation of Parts for Accurate 3D Object Detection and Tracking in Monocular Images. In *ICCV*, 2015. 1, 2, 3, 4

[4] A. Doumanoglou, V. Balntas, R. Kouskouridas, S. Malassiotis, and T. Kim. 6D Object Detection and Next-Best-View Prediction in the Crowd. In *CVPR*, 2016. 1

[5] B. Drost, M. Ulrich, N. Navab, and S. Ilic. Model Globally, Match Locally: Efficient and Robust 3D Object Recognition. In *CVPR*, 2010. 1, 2

[6] S. Gupta, P. Arbelaez, R. Girshick, and J. Malik. Aligning 3D Models to RGB-D Images of Cluttered Scenes. In *CVPR*, 2015. 3

[7] S. Hinterstoisser, C. Cagniart, S. Ilic, P. Sturm, N. Navab, P. Fua, and V. Lepetit. Gradient Response Maps for Real-Time Detection of Textureless Objects. *PAMI*, 2012. 1, 2, 6

[8] S. Hinterstoisser, V. Lepetit, S. Ilic, S. Holzer, G. Bradski, K. Konolige, and N. Navab. Model Based Training, Detection and Pose Estimation of Texture-Less 3D Objects in Heavily Cluttered Scenes. In *ACCV*, 2012. 6

[9] S. Hinterstoisser, V. Lepetit, N. Rajkumar, and K. Konolige. Going Further with Point Pair Features. In *ECCV*, 2016. 1, 2

[10] T. Hodan, P. Haluza, S. Obdrzalek, J. Matas, M. Lourakis, and X. Zabulis. T-LESS: An RGB-D Dataset for 6D Pose Estimation of Texture-less Objects. In *IEEE Winter Conference on Applications of Computer Vision*, 2017. 1, 6, 7, 8

[11] W. Kehl, F. Tombari, N. Navab, S. Ilic, and V. Lepetit. Hashmod: A Hashing Method for Scalable 3D Object Detection. In *BMVC*, 2015. 2

[12] A. Kendall, M. Grimes, and R. Cipolla. Posenet: A Convolutional Network for Real-Time 6-DOF Camera Relocalization. In *ICCV*, 2015. 3, 4

[13] A. Krull, E. Brachmann, F. Michel, M. Y. Yang, S. Gumhold, and C. Rother. Learning Analysis-By-Synthesis for 6D Pose Estimation in RGB-D Images. In *ICCV*, 2015. 1, 2

[14] K. Lai, L. Bo, X. Ren, and D. Fox. A Scalable Tree-Based Approach for Joint Object and Pose Recognition. In *AAAI*, 2011. 1, 2

[15] J. Long, E. Shelhamer, and T. Darrell. Fully Convolutional Networks for Semantic Segmentation. In *CVPR*, 2015. 3

[16] D. Lowe. Distinctive Image Features from Scale-Invariant Keypoints. *IJCV*, 20(2), 2004. 2

[17] M. Oberweger, P. Wohlhart, and V. Lepetit. Training a Feedback Loop for Hand Pose Estimation. In *ICCV*, 2015. 2, 5

[18] O. Russakovsky, J. Deng, H. Su, J. Krause, S.Satheesh, S. Ma, Z. Huang, A. Karpathy, A. Khosla, M. Bernstein, A. Berg, and L. Fei-Fei. Imagenet Large Scale Visual Recognition Challenge. *IJCV*, 115(3):211–252, 2015. 6

[19] J. Shotton, B. Glocker, C. Zach, S. Izadi, A. Criminisi, and A. Fitzgibbon. Scene Coordinate Regression Forests for Camera Relocalization in RGB-D Images. In *CVPR*, 2013. 2, 6

[20] K. Simonyan and A. Zisserman. Very deep convolutional networks for large-scale image recognition. *CoRR*, abs/1409.1556, 2014. 3, 4

[21] A. Tejani, D. Tang, R. Kouskouridas, and T.-K. Kim. Latent-Class Hough Forests for 3D Object Detection and Pose Estimation. In *ECCV*, 2014. 1, 2, 3

[22] Z. Tu and X. Bai. Auto-Context and Its Applications to High-Level Vision Tasks and 3D Brain Image Segmentation. *PAMI*, 2009. 3

[23] D. Wagner, G. Reitmayr, A. Mulloni, T. Drummond, and D. Schmalstieg. Pose Tracking from Natural Features on Mobile Phones. In *ISMAR*, 2008. 2

[24] P. Wohlhart and V. Lepetit. Learning Descriptors for Object Recognition and 3D Pose Estimation. In *CVPR*, 2015. 2